\documentclass[twoside]{article}
\usepackage[accepted]{aistats2019}
\usepackage[numbers]{natbib}

\usepackage{hyperref}
\usepackage{url}
\usepackage{sidecap}

\usepackage[utf8x]{inputenc}
\usepackage[T1]{fontenc}

\usepackage{amsmath} 
\usepackage{graphicx}
\usepackage{algorithm}
\usepackage[noend]{algpseudocode}
\usepackage{booktabs} 
\usepackage{subcaption}

\usepackage{titlesec}
\titlelabel{\thetitle.\quad}
\date{}

\makeatletter
\def\BState{\State\hskip-\ALG@thistlm}
\makeatother

\usepackage[inline]{enumitem}

\usepackage{tikz}
\usetikzlibrary{shapes.geometric, arrows, fit}
\tikzstyle{io} = [trapezium, trapezium left angle=70, trapezium right angle=120, minimum width=2cm, minimum height=1cm, text centered, draw=black, fill={rgb:red,1;green,2;blue,5;white,6}, trapezium stretches=true]
\tikzstyle{process} = [rectangle, minimum width=2cm, minimum height=1cm, text centered, draw=black, fill={rgb:red,1;green,2;blue,5;white,20}]
\tikzstyle{decision} = [diamond, minimum width=3cm, minimum height=1cm, text centered, draw=black, fill={rgb:red,1;green,2;blue,5;white,1}]
\tikzstyle{arrow} = [thick,->,>=stealth]

\usepackage{xspace}

\newcommand{\ie}{i.e.,\xspace}

\newcommand{\Xf}{\textbf{X}}
\newcommand{\Pa}{\text{Pa}}

\usepackage[colorinlistoftodos]{todonotes}

\title{\textbf{Universal Marginalizer for Amortised Inference and Embedding of Generative Models}}

\author{
        \textbf{Robert Walecki*,
        Albert Buchard,
        Kostis Gourgoulias,
        Chris Hart,
        Maria Lomeli,
        }\\
        \textbf{
        A. K. W. Navarro,
        Max Zwiessele,
        Yura Perov,
        Saurabh Johri*}
        \\
Babylon Health, London, UK\\
*\textbf{e-mail:} \textit{\{robert.walecki,saurabh.johri\}@babylonhealth.com}
}

\def\Pa{\mathrm{pa}}

\def\XX{\mathbf{X}}
\def\XO{\mathbf{X}_{\mathcal{O}}}
\def\xO{\mathbf{x}_{\mathcal{O}}}
\def\xxO{\mathbf{\tilde x}_{\mathcal{O}}}
\def\xxS{\mathbf{\tilde x}_{\mathcal{S}}}
\def\XU{\mathbf{X}_{\mathcal{U}}}
\def\xU{\mathbf{x}_{\mathcal{U}}}

\def\xS{\mathbf{x}_{\mathcal{S}}}

\def\xxSuO{\mathbf{\tilde x}_{\mathcal{S}\cup\mathcal{O}}}

\def\xSnPai {\mathbf{x}_{\mathcal{S}\cap\Pa(X_i)}}

\def\UM{\mathrm{UM}}

\begin{document}
\maketitle
\begin{abstract}
Probabilistic graphical models are powerful tools which allow us to formalise our knowledge about the world and
reason about its inherent uncertainty. There exist a considerable number of methods for performing inference in probabilistic graphical models; however, they can be computationally costly due to significant time burden and/or storage requirements; or they lack theoretical guarantees of convergence and accuracy when applied to large scale graphical models. 
To this end, we propose the Universal Marginaliser Importance Sampler (UM-IS) -- a hybrid inference scheme that combines the flexibility of a deep neural network trained on samples from the model and inherits the asymptotic guarantees of importance sampling. 
We show how combining samples drawn from the graphical model with an appropriate masking function allows us to train a single neural network to approximate any of the corresponding
conditional marginal distributions, and thus amortise the cost of inference. 
We also show that the graph embeddings can be applied for tasks such as: clustering, classification and interpretation of relationships between the nodes.
Finally, we benchmark the method on a large graph (>1000 nodes), showing that UM-IS outperforms sampling-based methods by a large margin while being computationally efficient.
\end{abstract}

\section{Introduction}
\label{sec:introduction}

Probabilistic Graphical Models (PGM) provide a natural framework for expressing the conditional independence relationships between random variables. PGMs are used to formalise our knowledge about the world and for reasoning and decision-making. 
PGMs have been successfully used for problems in a wide range of real-life applications including information technology, engineering, systems biology and medicine, among others.
In systems biology, the structure of a PGM is usualy learned and used to infer different biological properties from data \citep{friedman2004inferring}. 
For this type of application, the structure (edges) of the network is the main output. 
A particular example of a PGM is a Bayesian Network (BN) where all variables in the graphical model are discrete.
These types of networks are widely used for medical applications like diagnosis systems. For this, the network structure is designed by experts and is then used to infer the conditional marginal probability of diseases given a set of evidence that contains observations for risk factors and/or symptoms (see Fig.~\ref{fig:results}).
In such domains, the penalty for errors during inference can be potentially life-threatening. This risk can be mitigated by choosing a more complex model for the underlying process. However, exact inference is often computationally intractable for complex models, and so approximate inference is required. Furthermore, if we increase the complexity of the models, then the cost of inference will increase accordingly, limiting the feasibility of available algorithms. Some approximate inference methods are: variational inference~\citep{wainwright2008graphical} and Monte Carlo methods such as importance sampling~\citep{neal2001annealed}. Variational inference methods can be fast but do not target the true posterior. Monte Carlo inference is consistent, but can be computationally expensive. 

In this paper, we propose the Universal Marginaliser Importance Sampler (UM-IS), an amortised inference-based method for graph representation and efficient computation of asymptotically exact marginals.
In order to compute the marginals, the UM still relies on Importance Sampling (IS). We use a guiding framework based on amortised inference that significantly improves the performance of the sampling algorithm rather than computing marginals from scratch every time we run the inference algorithm.
This speed-up allows us to apply our inference scheme on large PGMs for interactive applications with minimum errors.
Furthermore, the neural network can be used to calculate a vectorised representation of the evidence nodes. This representations can then be used for various machine learning tasks such as node clustering and classification.

The main contributions of the proposed work are as follows:
\begin{itemize}
        \item We introduce UM-IS, a novel algorithm for amortised inference-based importance sampling. The model has the flexibility of a deep neural network to perform amortised inference. The neural network is trained purely on samples from the model prior and it benfits from the asymptotic guarantees of importance sampling.
        \item We demonstrate that the efficiency of importance sampling is significantly improved, which makes the proposed method applicable for interactive applications that rely on large PGMs.
        \item We show on a variety of toy network and on a medical knowledge graph (>1000 nodes) that the proposed UM-IS outperforms sampling-based and deep learning-based methods by a large margin, while being computational efficient.
        \item We show that the networks embeddings can serve as a vectorised representation of the provided evidence for tasks like classification and clustering or interpretation of node relationships.
\end{itemize}

\begin{figure*}[t!]
\centering
\includegraphics[width=1\linewidth]{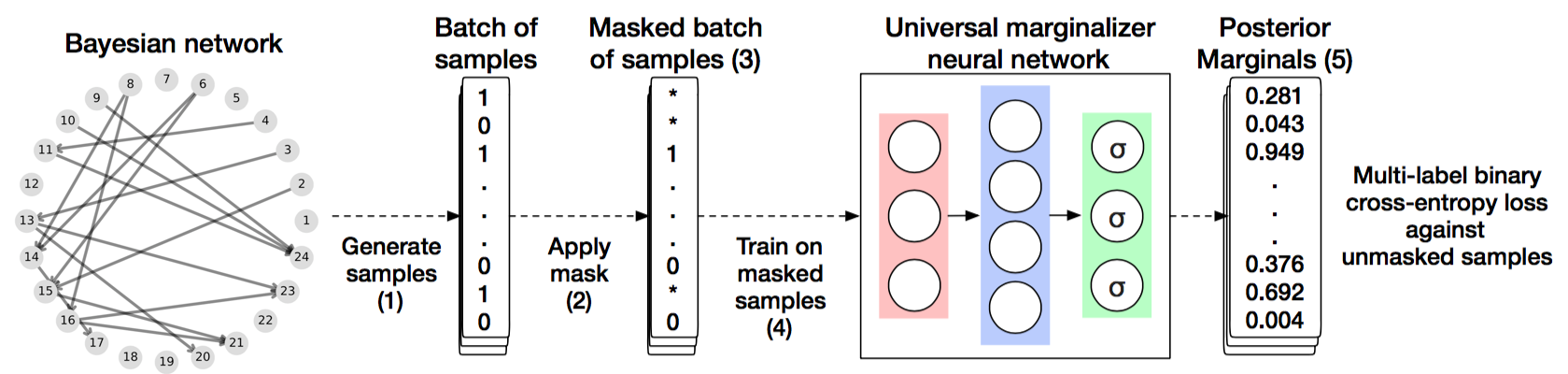}
\label{fig:pipeline}
        \caption{Universal Marginaliser: The UM performs scalable and efficient inference on graphical models. This figure shows one pass through the network. First, (1) a sample is drawn from the PGM, (2) values are then masked and (3) the masked set is passed through the UM, which then, (4) computes the marginal posteriors.}
\end{figure*}

\section{Related Work}
Currently, inference schemes in general PGMs use either message passing algorithms~\citep{murphy1999loopy}, variational inference~\citep{wainwright2008graphical,ng2000approximate,jordan1999introduction,jaakkola1999variational} or Markov Chain Monte Carlo \citep{hastings1970monte}.
Some exact inference algorithms are computationally expensive, within the context of the junction tree construction, because the time complexity is exponential
in the size of the maximal clique in the junction tree~\citep{jordan1999introduction}.
In some cases, exact methods can be computationally efficient in a small graph or sparse regime~\citep{heckerman1990tractable}. However, it has been shown that on larger graphs such methods converge to a local minimum \citep{heskes2003stable} that can be very different from the real marginals.
Importance sampling methods~\citep{cheng2000ais,neal2001annealed} are well studied and converge asymptotically to the global optimum. The caveat is that constructing good importance sampling proposals for large PGMs is hard and requires expert knowledge~\citep{shwe1991empirical}. For this reason, we focus on \textit{amortised inference}, techniques which speed up sampling by allowing us to ``flexibly reuse inferences so as to answer a variety of related queries''~\citep{gershman2014amortized}.

Amortised inference has been popular for Sequential Monte Carlo and has been used to learn in advance either parameters~\citep{gu2015neural} or a discriminative model which provides conditional density estimates~\citep{Morris:2001:RNA:2074022.2074068, paige2016inference}. 
These conditional density estimates can be used as proposals for importance sampling. This approach was also explored in~\citep{le2017using}. 
The authors use MADE, a fixed sequential density estimator~\citep{germain2015made}. 
In contrast, our method can be seen as further extension of MADE, a general density estimator, able to learn from arbitrary sets of evidence.

Feed-forward neural networks have recently been deployed to perform amortised inference~\citep{mnih2014neural, rezende2014stochastic}. 
For this application, neural networks are serving as non-iterative approximate inference methods, trained by minimising the error between sets of evidence and predicted posteriors. They have been successfully applied to a variety of computer vision tasks, where the graphical model and its corresponding neural network for inference is trained jointly by maximising the variational evidence lower bound~\citep{mnih2014neural}.
In a similar fashion, \cite{rezende2014stochastic} introduced stochastic back-propagation, a set of rules for gradient back-propagation through stochastic variables. The algorithm can be used to perform highly efficient inference in large scale PGMs.

Recently, probabilistic programming languages have become popular for describing and performing inference in a variety of PGMs bypassing the burden on the user of having to implement the inference method.
For example, \cite{ritchie2016deep} applied deep amortised inference to learn network parameters and later perform approximate inference on a PGM.
Such models either follow the control flow of a predefined sequential procedure, or are restricted to a fixed set of evidence. 





\section{Universal Marginalizer (UM)}
\label{section:um}
The Universal Marginaliser (UM) is a feed-forward neural network, used to perform fast, single-pass approximate inference on general PGMs at any scale. The UM can be used together with importance sampling as the proposal distribution, to obtain asymptotically exact results when estimating marginals of interest. We refer to this hybrid model as the Universal Marginaliser Importance Sampler (UM-IS). In this section, we introduce the notation and the training algorithm for the UM
(see supplementary material Section 1 for an introduction to importance sampling).
\subsection{Notation}
A Bayesian Network (BN) encodes a distribution $P$ over the random variables $\Xf=\{X_1,\ldots, X_N\}$ through a Directed Acyclic Graph (DAG), the random variables are the graph nodes and the edges dictate the conditional independence relationships between random variables. Specifically, the conditional independence of a random variable $X_i$ given its parents $\Pa(X_i)$ is denoted as $P(X_i|\Pa(X_i))$.

The random variables can be divided into two disjoint sets, $\XO \subset \Xf$ the set of observed variables within the BN, and $\XU \subset \Xf\setminus\XO$ the set of the unobserved variables.

We utilise a Neural Network (NN) as an approximation to the marginal posterior distributions $P(X_i| \XO = \xO)$ for each variable $X_i \in \XX$ given an instantiation $\xO$ of \emph{any} set of observations. We define $\xxO$ as the encoding of the instantiation that specifies which variables are observed, and what their values are (see Section~\ref{sec:methods}). For a set of binary variables $X_i$ with $i \in {0, ..., N}$, the desired network maps the $N$-dimensional binary vector $\xxO \subset B^N$ to a vector in $[0,1]^N$ representing the probabilities $p_i:=P(X_i= 1 | \XO = \xO)$:
\begin{align}
        \label{eq:umy}
  \mathbf{Y}=\UM(\xxO) \approx (p_1,\ldots, p_N).
\end{align}
This NN is used as a function approximator, hence, it can approximate any posterior marginal distribution given an arbitrary set of evidence $\XO$. For this reason, we call this discriminative model as the \textit{Universal Marginaliser} (UM). Indeed, if we consider the marginalisation operation in a Bayesian Network as a function $f:B^N\rightarrow[0,1]^N$, then existence of a neural network which can approximate this function is a direct consequence of the Universal Function Approximation Theorem (UFAT)~\citep{hornik}. It states that, under mild assumptions of smoothness, any continuous function can be approximated to an arbitrary precision by a neural network of a finite, but sufficiently large, number of hidden units. 
Once the weights of the NN are optimised, the activations of those hidden units can be computed to any new set of evidence. They are a compressed vectorised representation of the evidence set and can be used for tasks such as node clustering or classification.

\subsection{Training a UM}
In this section, we describe each step of the UM's training algorithm for a given PGM. This model is typically a multi-output NN with one output per node in the PGM (i.e. each variable $X_i$).
Once trained, this model can handle any type of input evidence instantiation and produce approximate posterior marginals $P(X_i = 1 | X_O = \xO)$.

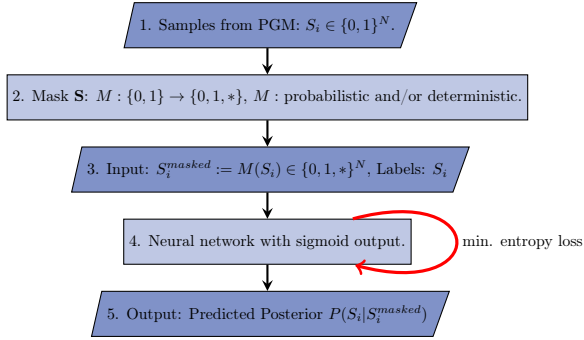
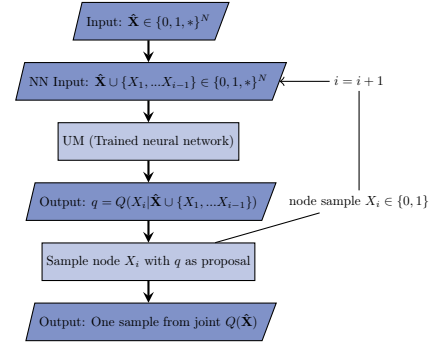
\begin{figure*}[ht]
\begin{subfigure}{0.4\linewidth}
  \centering
  \begin{tikzpicture}[node distance=1.6cm, scale=0.60, transform shape]
    \node (PGM) [io] {1. Samples from PGM: $S_i\in \{0,1\}^N$.}; 
    \node (mask) [process, below of=PGM] {2. Mask $\textbf{S}$: $M:\{0,1\}\to \{0,1,*\}$,
      $M:$ probabilistic and/or deterministic.}; 
    \node (masked) [io, below of=mask] {3. Input: $S^{masked}_i := M(S_i)\in\{0,1,*\}^N$, Labels: $S_i$}; 
    \draw [arrow] (PGM) -- (mask);  
    \node (train) [process, below of=masked] {4. Neural network with sigmoid output.};
        \draw [arrow] (mask) -- (masked);  
          \node (output) [io, below of=train] {5. Output: Predicted
      Posterior $P(S_i | S^{masked}_i)$}; 
      \draw [arrow] (train) -- (output); \draw [arrow] (masked) -- (train);
    \path [->, red, very thick] (train) edge[loop right] node (CE)
    [right, black] {min. entropy loss} (train) ;
  \end{tikzpicture}
  \caption{\textbf{UM Training:} The process to train a Universal Marginaliser using binary data generated from a Bayesian Network}
  \label{fig:train}
\end{subfigure}
\hfill
\begin{subfigure}{0.4\linewidth}
  \centering
  \begin{tikzpicture}[node distance=1.6cm, scale=0.5, transform shape]
    \node (inp) [io] {Input: $\mathbf{\hat{X}}\in\{0,1,*\}^N$};
    \node (test_data) [io, below of=inp] {NN Input: $\mathbf{\hat{X}} \cup \{X_1,...X_{i-1}\} \in\{0,1,*\}^N$};
    \draw [arrow] (inp) -- (test_data);
          \node (train_output) [process, below of=test_data] {UM (Trained neural network)}; 
    \draw [arrow] (test_data) -- (train_output); 
    \node (sample) [io, below of=train_output] {Output: $q =
      Q(X_i | \mathbf{\hat{X}} \cup \{X_1,...X_{i-1}\})$}; \node (impo) [process, below
    of=sample] {Sample node $X_i$ with $q$ as proposal}; \draw
    [arrow] (sample) -- (impo); 
    \draw [arrow] (train_output) -- (sample); 
    \node (i) [right of=test_data,xshift=4cm] {$i = i+1$}; 
    \node (ce) [right of=sample,xshift=4cm] {node sample $X_i \in \{0,1\}$}; \draw [->, black] (impo) -- (ce) -- (i) -- (test_data) ; 
    \node (full_sample) [io, below of=impo] {Output: One sample from joint $Q(\mathbf{\hat{X}})$}; \draw [arrow] (impo) -- (full_sample); 
  \end{tikzpicture}
        \caption{\textbf{Inference using UM-IS}: The part in the box is repeated $N$ times, for each node $i$ in topological order}
  \label{fig:inf}
\end{subfigure}
    \caption{Training and inference of the UM-IS.}
\end{figure*}

The flow chart with each step of the training algorithm is depicted in Fig.~\ref{fig:train}.
For simplicity, we assume that the training data (samples for the PGM) is pre-computed, and only one epoch is used to train the UM.

In practice, the following steps 1--4 are applied for each of the mini-batches separately rather than on a full training set all at once.
This improves memory efficiency during training and ensures that the network receives a large variety of evidence combinations, accounting for low probability regions in $P$. The steps are given as follows:

\noindent {\bf 1.\ Acquiring samples from the PGM.}
The UM is trained offline by generating unbiased samples (\ie complete assignment) from the PGM using ancestral sampling \cite[Algorithm 12.2]{koller2009probabilistic}. The PGM described here only contains binary variables $X_i$, and each sample $S_i \in B^N$ is a binary vector. In the next steps, these vectors will be partially masked as input and the UM will be trained to reconstruct the complete unmasked vectors as output.

\noindent {\bf 2.\ Masking.}
In order for the network to approximate the marginal posteriors at test time, and be able to do so for any input evidence, each sample $S_i$ must be partially masked. The network will then receive as input a binary vector where a subset of the nodes initially observed were hidden, or \textit{masked}. This masking can be deterministic, \ie always masking specific nodes, or probabilistic. We use a different masking distribution for every iteration during the optimization process. This is achieved in two steps. First, we sample two random numbers from a uniform distribution $i,j \sim U[0,N]$ where N is the number of nodes in the graph. Next, we mask from randomly selected $i$ ($j$) number of nodes the positive (negative) state.
In this way, the ratio between the positive and negative evidence and the total number of masked nodes is different with every iteration. A network with a large enough capacity will eventually learn to capture all these possible representations.

There is some analogy here to dropout in the input layer and so this approach could work well as a regulariser, independently of this problem \citep{JMLR:v15:srivastava14a}. However, it is not suitable for this problem because of the constant dropout probability for all nodes.

\noindent {\bf 3.\ Encoding the masked elements.}
Masked elements in the input vectors $S^{masked}_{i}$ artificially reproduce queries with unobserved variables, and so their encoding must be consistent with the one used at test time. The encodings are detailed in Section~\ref{sec:methods}.

\noindent {\bf 4.\ Training with Cross Entropy Loss.}
We trained the NN by minimising the multi-label binary cross entropy of the sigmoid output layer and the unmasked samples $S_i$.

\noindent {\bf 5.\ Outputs: Posterior marginals.}
The desired posterior marginals are approximated by the output of the last NN-layer. We can use these values as a first estimate of the marginal posteriors (UM approach); however, combined with importance sampling, these approximated values can be further refined (UM-IS approach). This is discussed in Sections~\ref{section:sampling_joint} and is empirically verified in Section~\ref{sec:results}.

\section{Hybrid: UM-IS}
\label{section:um_is}
\subsection{Sequential UM for Importance Sampling}
\label{section:sampling_joint}
The UM is a discriminative model which, given a set of observations $\XO$, will approximate all the posterior marginals. 
While useful on its own, the estimated marginals are not guaranteed to be unbiased. To obtain a guarantee of asymptotic unbiasedness while making use of the speed of the approximate solution, we use the estimated marginals for proposals in importance sampling. A na\"{i}ve approach is to sample each $X_i \in \XU$ independently from $\UM(\xxO)_i$, where $\UM(\xxO)_i$ is the $i$-th element of vector $\UM(\xxO)$.  However, the product of the (approximate) posterior marginals may be very different to the true posterior joint, even if the marginal approximations are good (see supplementary material Section 2 for more details).

The universality of the UM makes the following scheme possible, which we call the \textit{Sequential Universal Marginaliser Importance Sampling} (SUM-IS). A single proposal is sampled $\xS$ sequentially as follows. First, a new partially observed state is introduced $\xxSuO$ and it is initialised to $\xxO$. Then, we sample $[\xS]_1 \sim \UM(\xxO)_1$, and update the previous sample $\xxSuO$ such that $X_1$ is now observed with this value.  We repeat this process, at each step sampling $[\xS]_i \sim \UM(\xxSuO)_i$, and updating $\xxSuO$ to include the new sampled value. Thus, we can approximate the conditional marginal for a node $i$ given the current sampled state $\mathbf{X_\mathcal{S}}$ and evidence $\XO$ to get the optimal proposal $Q_i^\star$ as follows:
\begin{align}
Q^\star_i &= P(X_i|\{X_1,\ldots, X_{i-1}\}\cup\mathbf{X_\mathcal{O}})
\approx \UM(\xxSuO)_i.
\end{align}
Thus, the full sample $\xS$ is drawn from an implicit encoding of the approximate posterior \emph{joint} distribution given by the UM. This is because the product of sampled probabilities from Equation~\ref{eq:topo} is expected to yield low variance importance weights when used as a proposal distribution.
\begin{align}
  \label{eq:topo}
  Q &= \UM(\xxO)_1\prod_{i=2}^{N}\UM(\xxSuO)_i\\
        &\approx P(X_1|\XO)\prod_{i=2}^{N}P(X_i|X_1,\ldots, X_{i-1},\XO).
\end{align}
The process by which we sample from these proposals is illustrated in Algorithm~\ref{algorithm:sumis} and in Fig.~\ref{fig:inf}.
\begin{algorithm}
\caption{Sequential Universal Marginalizer importance sampling}
\label{algorithm:sumis}
\begin{algorithmic}[1]
\State Order the nodes topologically $X_1,...X_N$, where $N$ is the total number of nodes.
\For{$j$ in [1,...,$M$] (where $M$ is the total number of samples):}
  \State $\xxS = \emptyset$
  \For{$i$ in [1,...$N$]:}
      \State sample node $x_i$ from $Q(X_i)=\UM(\xxSuO)_i \approx P(X_i |\mathbf{X_\mathcal{S}}, \mathbf{X_\mathcal{O}})$
      \State add $x_i$ to $\xxS$
  \EndFor
  \State $[\xS]_j = \xxS$
  \State $w_j = \prod_{i=1}^{N}\frac{P_i}{Q_i}$ (where $P_i$ is the likelihood, $P_i = P(X_i=x_i| \xSnPai)$ and $Q_i=Q(X_i=x_i)$)
\EndFor
\State $E_p[X] = \frac{\sum_{j=1}^{M}X_j w_j}{\sum_{j=1}^{M}w_j}$ (as in standard IS)
\end{algorithmic}
\end{algorithm}

The nodes are sampled sequentially, using the UM to provide a conditional probability estimate at each step. This requirement can affect computation time, depending on the parallelisation scheme used for sampling. In our experiments, we observed that some parallelisation efficiency can be recovered by increasing the number of samples per batch. 
\subsection{UM Architecture}
\begin{figure*}
\hspace{1cm}
\begin{subfigure}{0.25\linewidth}
        \includegraphics[width=\linewidth]{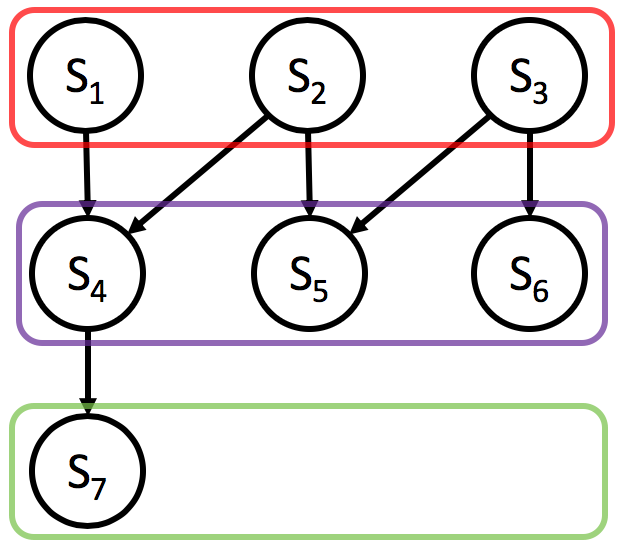}
        \caption{Directed Graphical Model}
\end{subfigure}
\hfill
\begin{subfigure}{0.50\linewidth}
        \includegraphics[width=\linewidth]{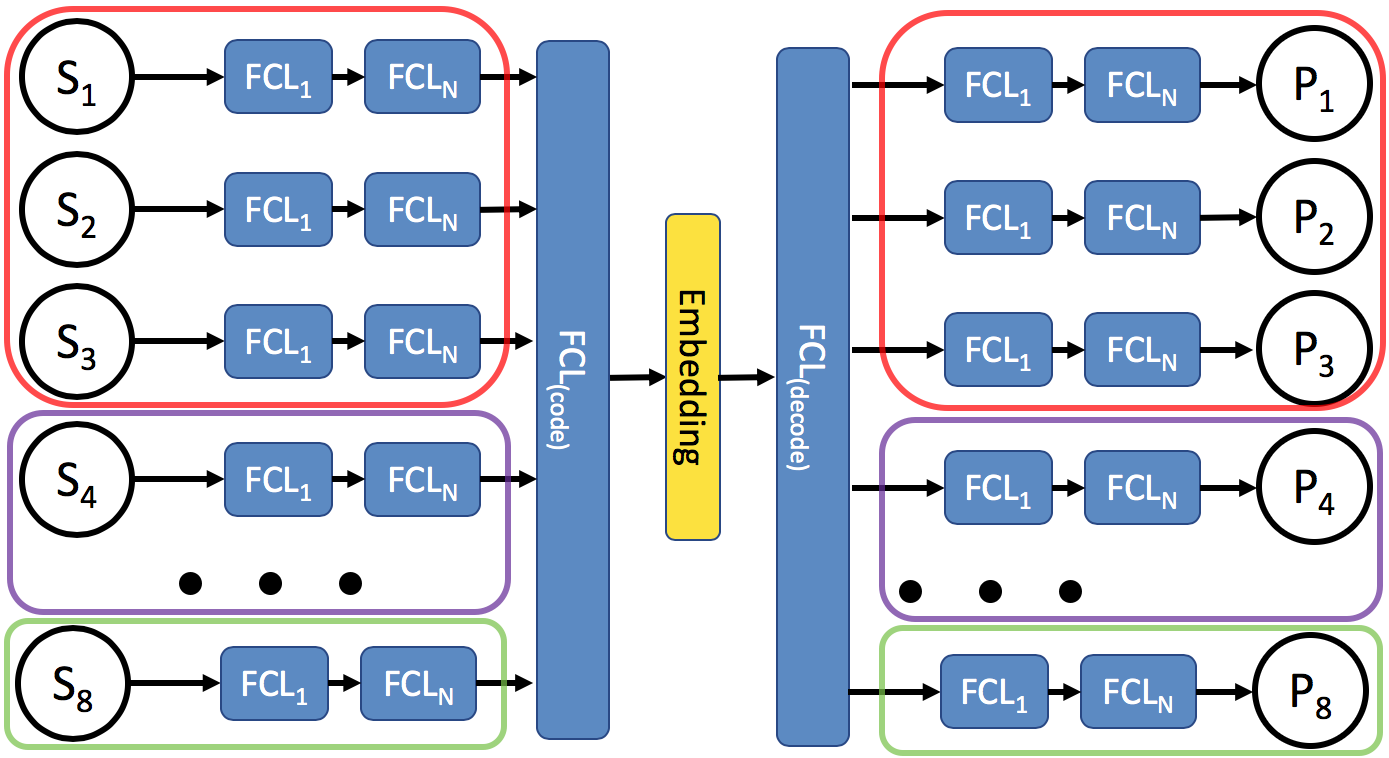}
        \caption{UM architecture}
\end{subfigure}
\hspace{1cm}
        \caption{Graphical Model and the corresponding UM architecture. The nodes of (a) the graph are categorized by their depth inside the network and the weights of (b) the UM neural network are shared for nodes of the same category.}
\label{fig:architecture}
\end{figure*}

The architecture of the UM is shown in Fig.~\ref{fig:architecture}. It is mostly similar to a denoising auto-encoder (see \cite{vincent2008extracting}) but with multiple branches -- one branch for each node of the graph.
In our experiments, we noticed that the cross entropy loss for different nodes highly depends on the number of parents and its depth in the graph. 
To simplify the network and reduce the number of parameters, we share the weights of all fully connected layers that correspond to specific type of nodes. The types are defined by the depth in the graph (type 1 nodes have no parents, type 2 nodes have only type 1 nodes as parents etc.). 
The architecture of the best performing model on the large medical graph has three types of nodes and the embedding layer has 2048 hidden states (more details are in Section \ref{sec:methods}).

\section{Experiments}
\label{sec:experiments}
In our experiments, we chose the best performing UM in terms of Mean Absolute Error (MAE) on the test set for the subsequent experiments.
We use ReLU non-linearities, apply dropout ~\citep{JMLR:v15:srivastava14a, baldi2014dropout} on the last hidden layer and use the Adam optimization method~\citep{kingma2014adam} with batchsize of 2000 samples per batch for parameter learning. We have also included batch normalization between the fully connected layers.
To train the model on a large medical graphical model, we used in total a stream of $3\times10^{11}$ samples, which took approximately 6 days on a single GPU. 

\subsection{Setup}
\label{sec:methods}
\textbf{Graph:} We carry out our experiments on a large (>1000 nodes) proprietary Bayesian Network for medical diagnosis representing the relationships between risk factors, diseases and symptoms. A illustration of the model structure is given in Fig.~\ref{fig:results_pgm}.\\
\\
\textbf{Model:} We tried different NN architectures with a grid search over the values of the hyperparameters and on the number of hidden layers, number of states per hidden layer, learning rate and strength of regularisation through dropout.\\
\\
\textbf{Test set:} The quality of approximate conditional marginals was measured using a test set of posterior marginals computed for 200 sets of evidence via ancestral sampling with 300 million samples. 
The test evidence set for the medical graph was generated by experts from real data. The test evidence set for the synthetic graphs was sampled from a uniform distribution. We used standard importance sampling, which corresponds to the likelihood weighting algorithm for discrete Bayesian networks~\citep[Chapter 12]{koller2009probabilistic}, with 8 GPUs over the course of 5 days to compute precise marginal posteriors of all test sets.\\
\\
\textbf{Metrics:} Two main metrics are considered: the Mean Absolute Error (MAE) given by the absolute difference of the true and predicted node posteriors and the Pearson Correlation Coefficient (PCC) of the true and predicted marginal vectors.
Note that we did not observe negative correlations and therefore both measures are bounded between 0 and 1.
We also used the Effective Sample Size (ESS) statistic for the comparison with the standard importance sampling. This statistics measures the efficiency of the different proposal distributions used during sampling. In this case, we do not have access to the normalising constant of the posterior distribution, the ESS is defined as $(\sum_{i=1}^{M}w_i)^2/\sum_{i=1}^{M}(w_i)^2$, where the weights, $w_i$, are defined in Step~8 of Algorithm~\ref{algorithm:sumis}.\\
\\
\textbf{Data Representation:} We consider a one hot encoding for the unobserved and observed nodes.
This representation only requires two binary values per node. One value represents if the node is observed and positive ([0,1]) and the other value represents whether this node is observed and negative ([1,0]). If the node is unobserved or masked, then both values are set to zero ([0,0]).
\subsection{Results}
\label{sec:results}
In this section, we first discuss the results of different architectures for the UM, then compare the performance of importance sampling with different proposal functions. Finally, we discuss the efficiency of the algorithm.
\newcommand{\WW}{0.3}
\begin{figure*}[ht!]
\centering
\begin{subfigure}{\WW\linewidth}
        \includegraphics[width=\linewidth]{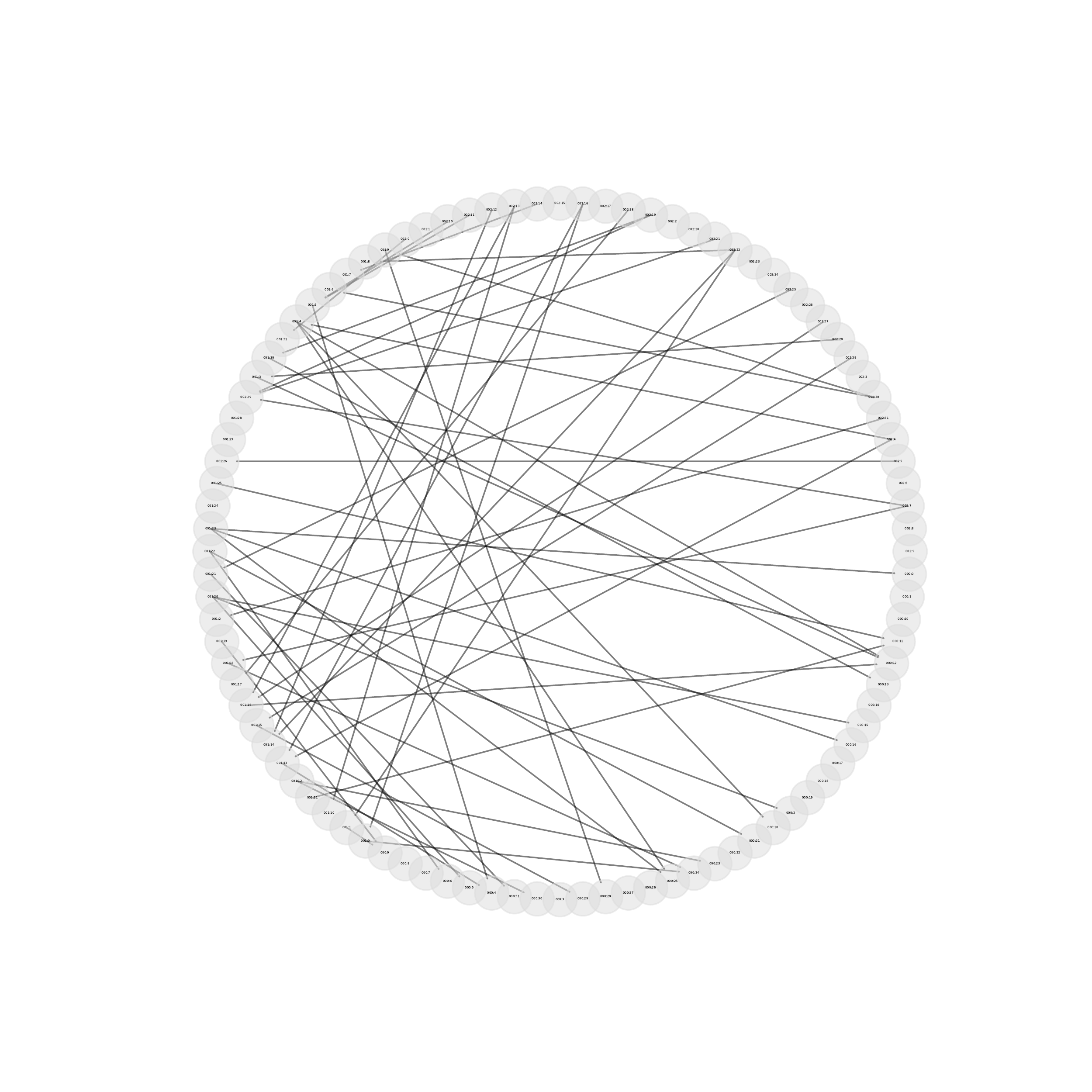}
        \includegraphics[width=\linewidth]{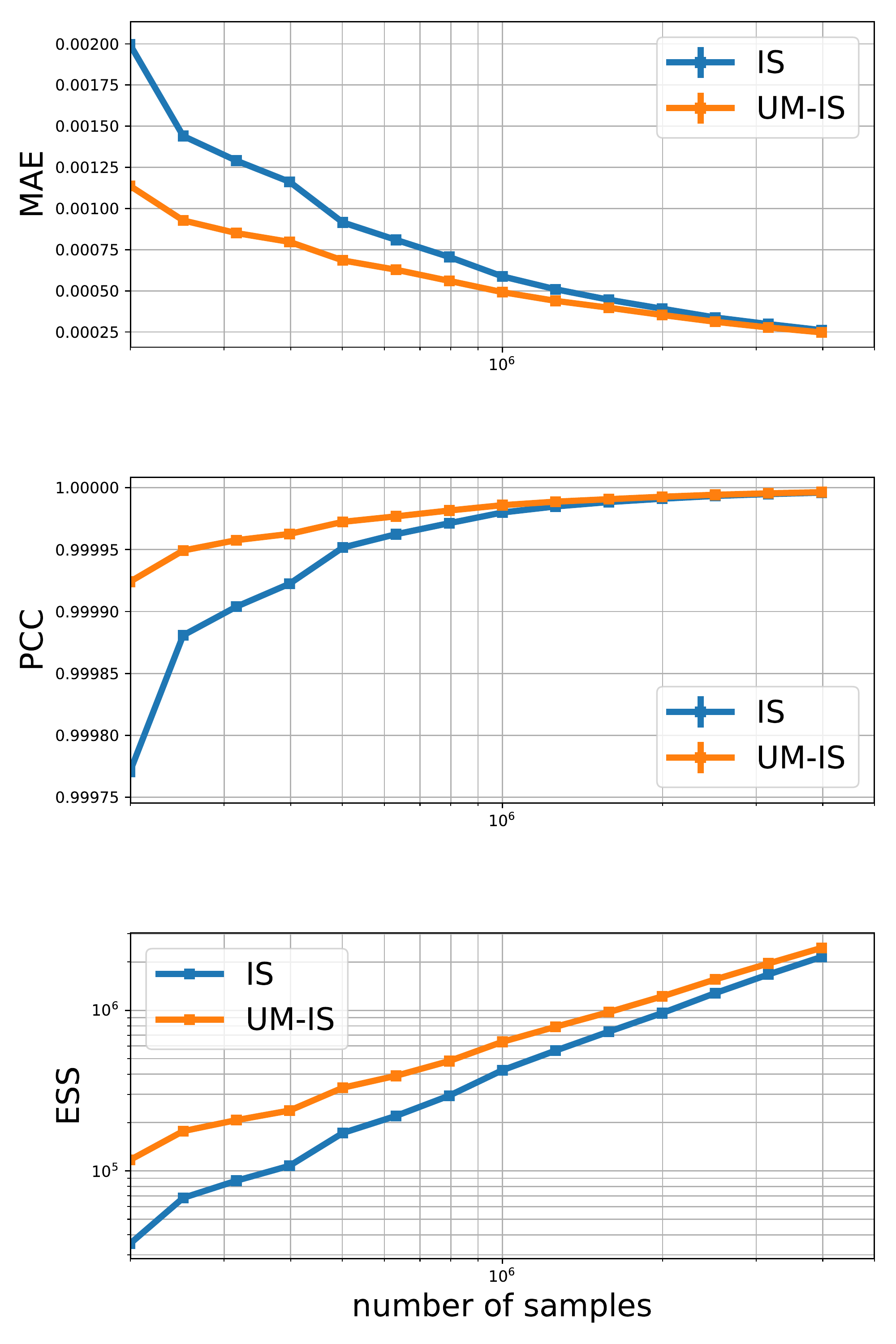}
        \caption{Synthetic graph, 96 nodes.}
        \label{fig:results_a}
\end{subfigure}
\begin{subfigure}{\WW\linewidth}
        \includegraphics[width=\linewidth]{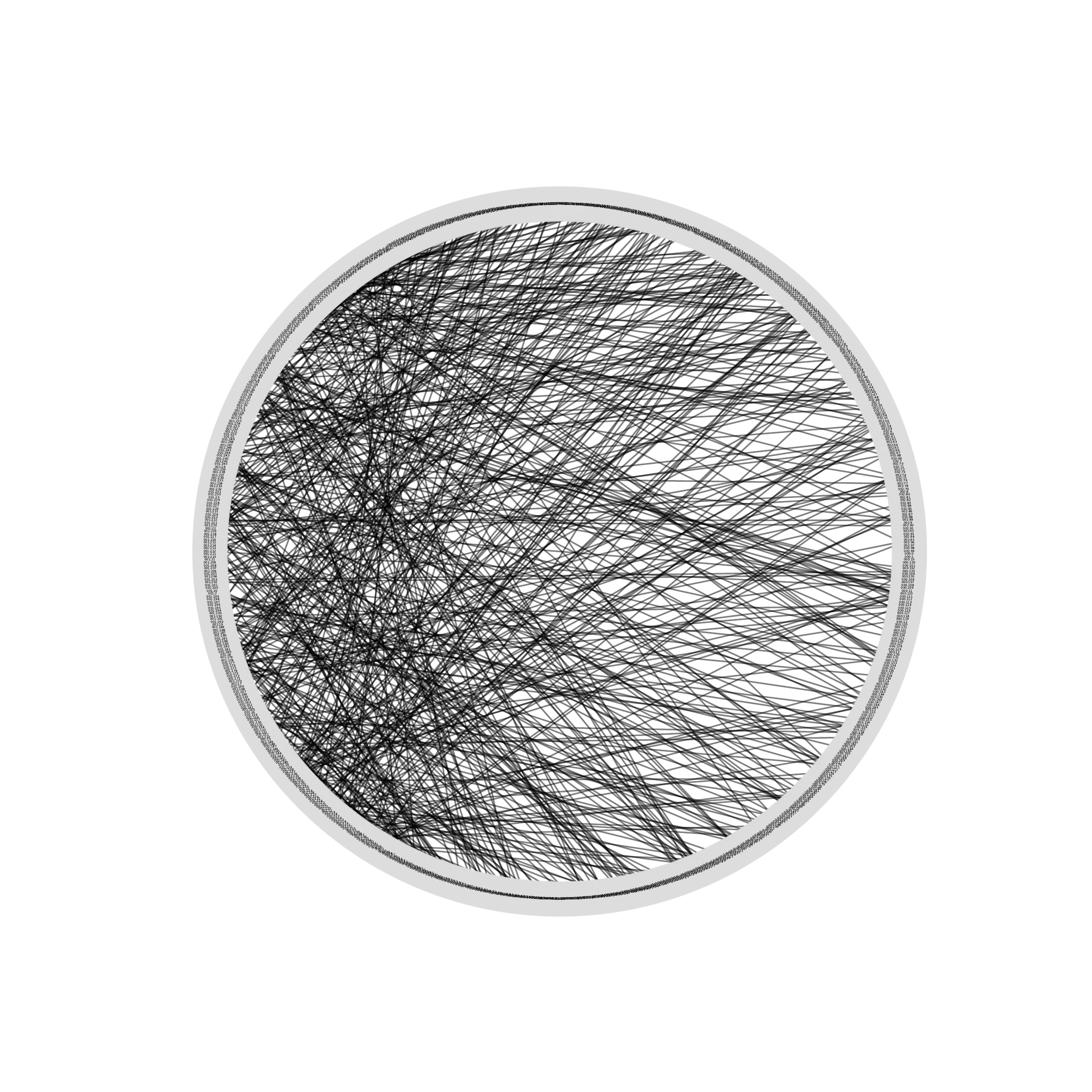}
        \includegraphics[width=\linewidth]{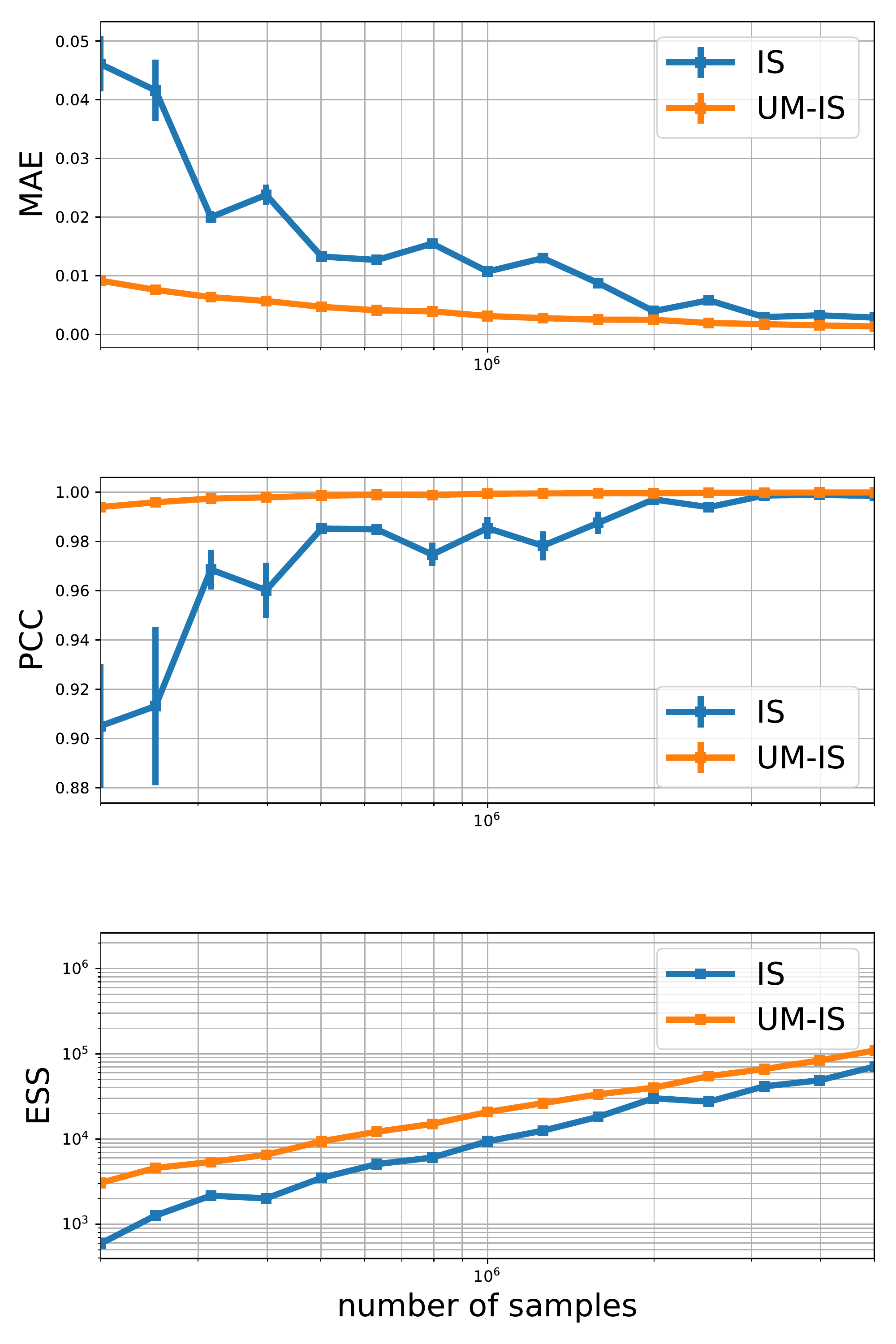}
        \caption{Synthetic graph, 768 nodes.}
        \label{fig:results_b}
\end{subfigure}
\begin{subfigure}{\WW\linewidth}
        \includegraphics[width=\linewidth]{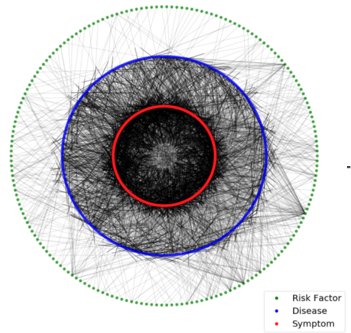}
        \includegraphics[width=\linewidth]{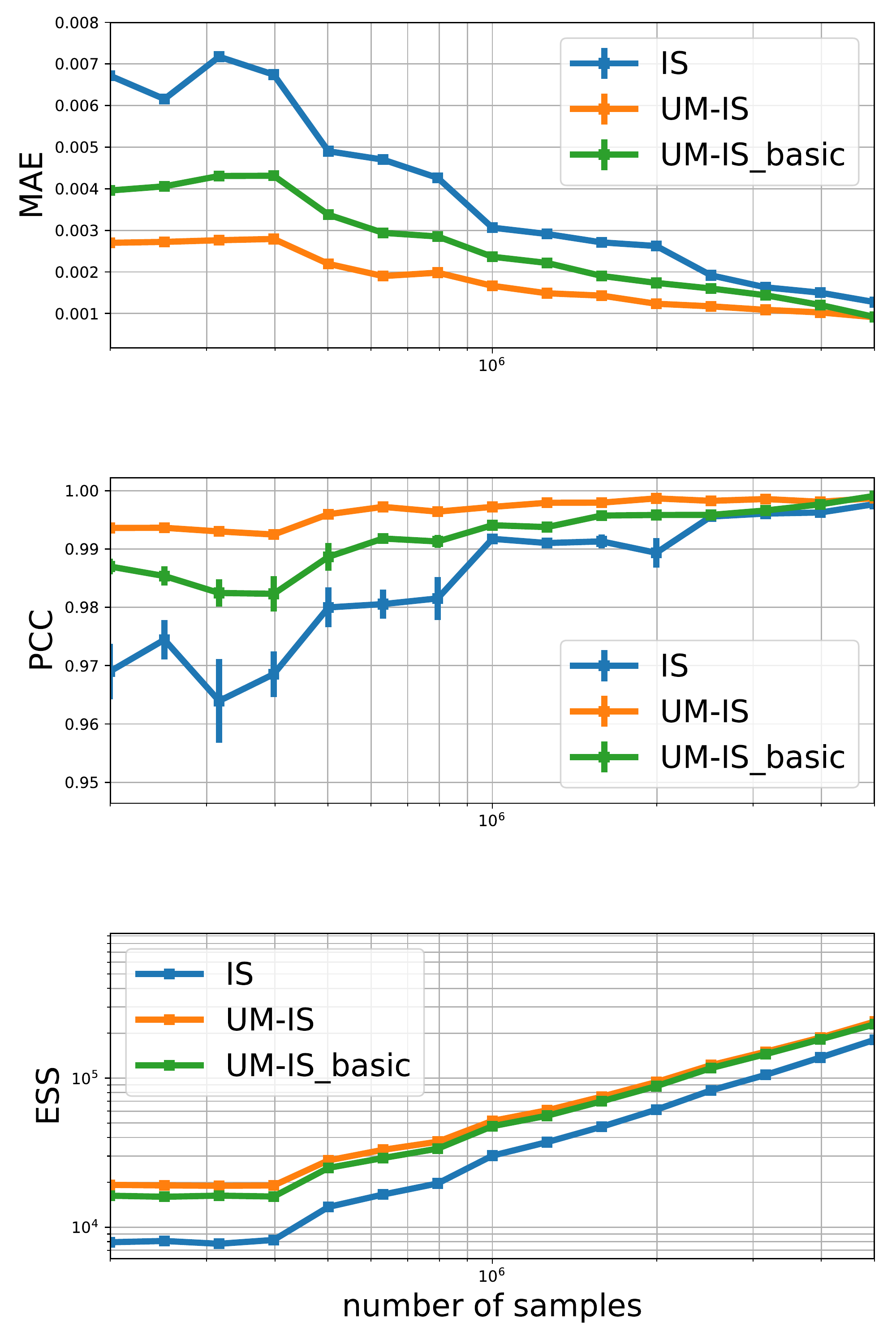}
        \caption{Medical PGM, ($\sim$1200 nodes.)}
        \label{fig:results_pgm}
\end{subfigure}
        \caption{Performance on three different graphical models. We applied inference through importance sampling with and without the support of a trained UM and evaluate it in terms of Pearson Correlation Coefficient (PCC), Mean Absolute Error (MAE) and Effective Sampling Size (ESS). The medical PGM described in the paper was designed with the help of the medical experts and contains $\sim$1200 nodes.}
\label{fig:results}
\end{figure*}

{\bf  UM Architecture and Performance:}
We used a hyperparameter grid search on the different network architectures and data representations.
The algorithmic performance was not greatly affected for different types of data representations. We hypothesise that this is due to the fact that neural networks are flexible models capable of handling different types of inputs efficiently by capturing the representations within the hidden layers.
In contrast, the network architecture of the UM strongly depends on the structure of the PGM. For this reason, a specific UM needs to be trained for each PGM. This task can be computationally expensive but once the UM is trained, it can be used to compute the approximate marginals in a single forward pass on any new and even unseen set of evidence. 
\\
\vspace{-2mm}

{\bf UM for Inference in PGMs:}
In order to evaluate the performance of sampling algorithms, we monitor the change in PCC and MAE on the test sets with respect to the total number of samples.
We notice that across all experiments, a faster increase in the  maximum value or the PCC is observed when the UM predictions are used as proposals for importance sampling. This effect becomes more pronounced as the size of the graphical model increases. 
Fig.~\ref{fig:results} indicates standard IS (blue line) reaches PCC close to 1 and an MAE close to 0 on the small network with 96 nodes. 
In this case of very small graphs, both algorithms converge quickly to the exact solution. However, UM-IS (orange-line) still outperforms IS and converges faster, as seen in Fig.~\ref{fig:results_a}.
For the synthetic graph with 798 nodes, standard IS reaches an MAE of $0.012$ with $10^6$ samples, whereas the UM-IS error is 3 times lower ($0.004$) for the same number of samples. The same conclusions can also be drawn for PCC.
Most interestingly, on the large medical PGM (Fig.~\ref{fig:results_pgm}), the UM-IS with $10^5$ samples exhibits better performance than standard IS with $10^5$ samples in terms of MAE and PCC. 
\begin{figure*}
\begin{subfigure}{0.43\linewidth}
        \includegraphics[width=1\linewidth]{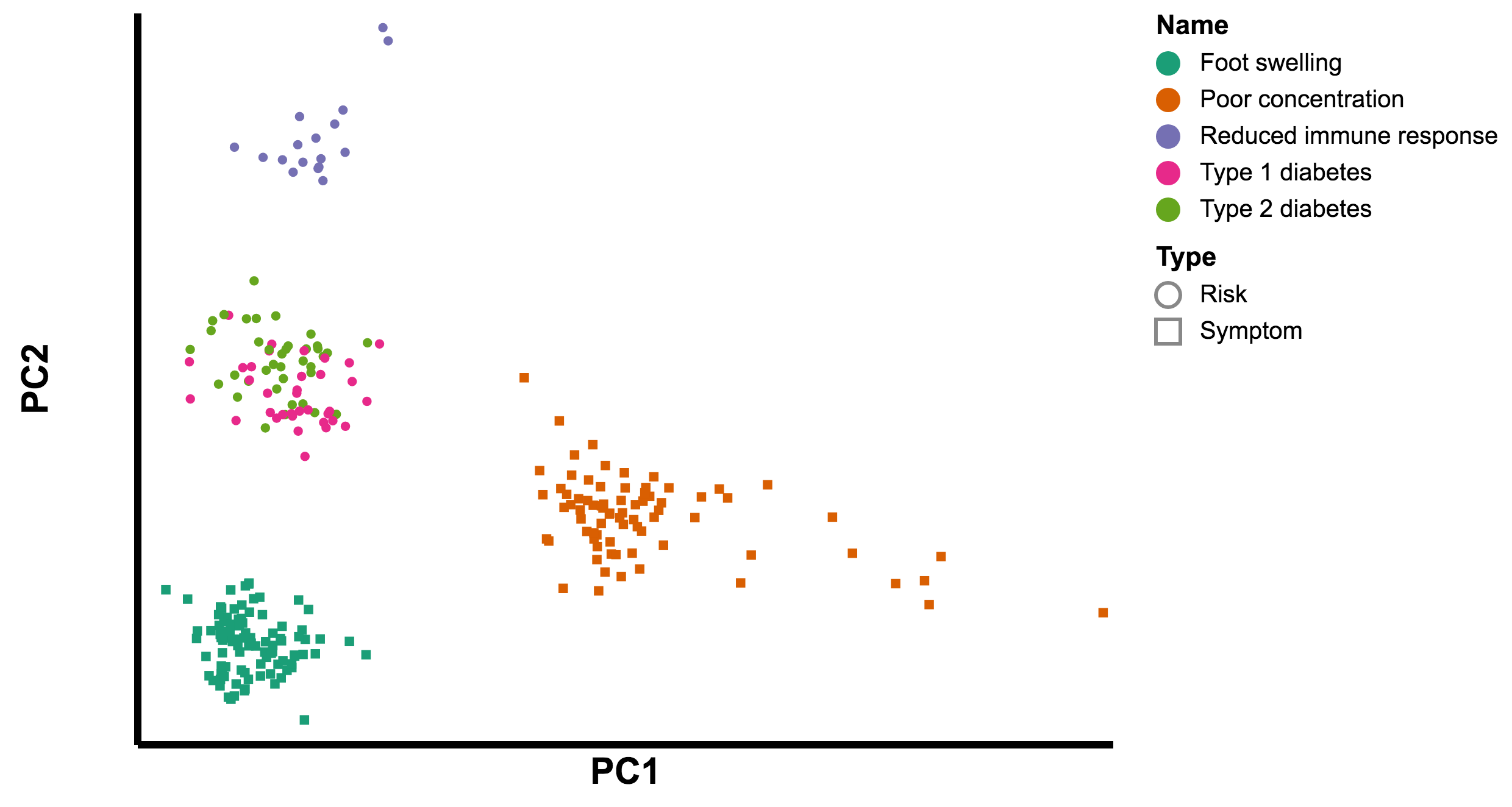}
        \caption{Diabetes embeddings.}
        \label{fig:dia}
\end{subfigure}
\hfill
\begin{subfigure}{0.43\linewidth}
        \includegraphics[width=1\linewidth]{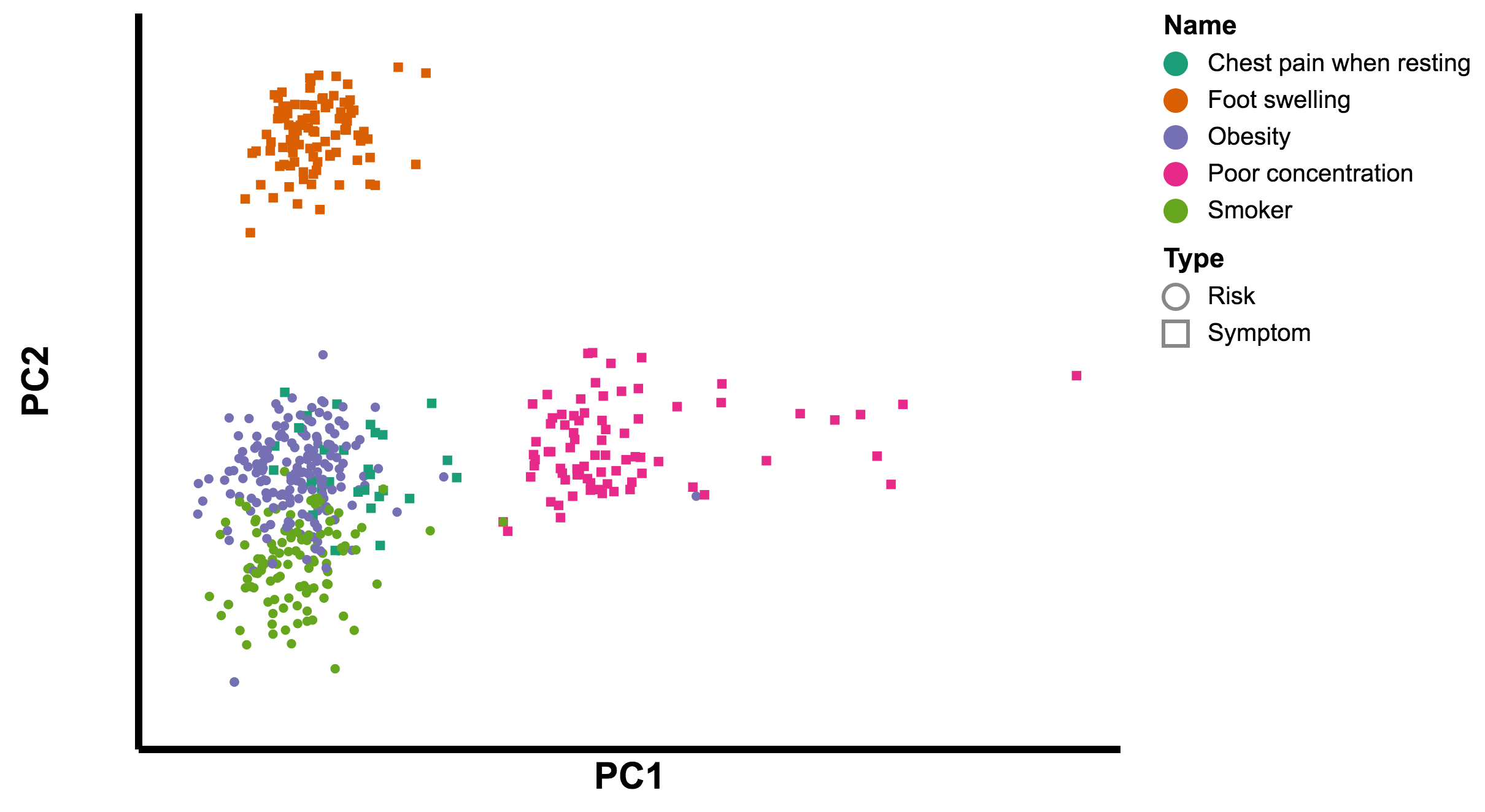}
        \caption{Smoke, Obesity embeddings.}
        \label{fig:smo}
\end{subfigure}
\caption{The figures show the embeddings filtered for two set of symptoms and risk factors, where each scatter point corresponds to a set of evidence. The display embedding vectors correspond to the first two components. It can be seen that they separate quite well unrelated medical concepts and show an overlap for concepts which are closely related.\label{fig:graph_embedding}}
\end{figure*}
In other words, the time (and computational costs) of the inference algorithm is significantly reduced by factor of ten or more.
We expect this improvement to be even stronger on much larger graphical models (see supplementary material Section 3 for more details).
We also include the results of a simple UM architecture as a baseline. This simple UM (UM-IS-Basic) has one single hidden layer that is shared for all nodes of the PGM.
We can see that the MAE and PCC still improved over standard IS. However, UM-IS with multiple fully connected layers per group of nodes significantly outperforms the basic UM by a large margin.
There are two reasons for this. First, the model capacity of the UM is higher which allows to learn more complex structures from the data. 
Secondly, the losses in the UM are spread across all groups of nodes and the gradient update steps are optimised with the right order of magnitude for each group.
This prevents the model from overfitting to the states of a specific type of node with a significant higher loss.
\\
\\
{\bf  Graph Embedding:}
Extracting meaningful representations form the evidence set is an additional interesting feature of the UM.
In this section, we demonstrate the qualitative results for this application.
The graph embeddings are extracted as the 2048 dimensional activations of the inner layer of the UM (see Fig.~\ref{fig:architecture}). 
They are a low-dimensional vectorised representation of the evidence set in which the graphs structure is preserved. That means that the distance for nodes that are tightly connected in the PGM should be smaller that the distance to nodes than are independent.
In order to visualise this feature, we plot the first two principal components of the embeddings from different evidence sets in which we know that they are related. We use the evidence set from the medical PGM with different diseases, risk-factors and symptoms as nodes.
Fig.~\ref{fig:dia} shows that the embeddings of sets with active Type-1 and Type-2 diabetes are collocated. Although the two diseases have different underlying cause and connections in the graphical model (i.e pancreatic beta-cell atrophy and insulin-resistance respectively), they share similar symptoms and complications (e.g cardiovascular diseases, neuropathy, increased risk of infections etc.).
A similar clustering can be seen in Fig.~\ref{fig:smo} for two cardiovascular risk factors: smoking and obesity, interestingly collocated with a sign seen in patient suffering from a severe heart condition (i.e unstable angina, or acute coronary syndrome): chest pain at rest.
\\
\\
{\bf  Node Classfication:}
\begin{table}
    \centering
        \tiny
        \caption{Classification performances using two different features. Each classifier is trained on - \textit{dense} the dense embedding as features, and \textit{input} - the top layer (UM input) as features.
    The target (output) is always the disease layer.}
    \begin{tabular}{lcccc}
    \toprule
    {}        & \multicolumn{2}{c}{Linear SVC}  & \multicolumn{2}{c}{Ridge}                                                                          \\
    {}        & dense                           & input                        & dense                           & input         \\
    \midrule
    F1        & $0.67\pm0.01$                   & $0.07\pm0.00$                & $0.66\pm0.04$                   & $0.17\pm0.01$ \\
    Precision & $0.84\pm0.03$                   & $0.20\pm0.04$                & $0.81\pm0.06$                   & $0.22\pm0.04$ \\
    Recall    & $0.58\pm0.02$                   & $0.05\pm0.00$                & $0.59\pm0.04$                   & $0.16\pm0.01$ \\
    Accuracy  & $0.69\pm0.01$                   & $0.31\pm0.01$                & $0.63\pm0.02$                   & $0.27\pm0.01$ \\
    \bottomrule
    \end{tabular}
    \label{tab:classification_embedding}
\end{table}
To further asses the quality of the UM embeddings, we performed experiments for node classification with different features and two different classifiers.
More precisely, we train a SVM and Ridge regression model with thresholded binary output for multi-task disease detection. 
These models were trained to detect the $14$ most frequent diseases from (a) the set of evidence or (b) the embedding of that set. 
We used 5-fold standard cross validation with a grid search over the hyperparameter of both models and the number of PCA components for data preprocessing.
Table \ref{tab:classification_embedding} shows the experimental results for the two types of features.
As expected, the models that were trained on the UM embeddings reach a significantly higher performance across all evaluation meassures. This is mainly because the embeddings of the evidence set are effectively compressed and structured and also preserve the information form the graph structure.
Note that the mapping from the evidence set to the embeddings was optimised with an large number of generated samples ($3*10^11$) during the UM learning phase. 
Therefore, these representations can be used to build more robust machine learning methods for classfication and clustering rather then using the raw evidence set to the PGM.

\section{Conclusion}
\label{sec:conclusion}
This paper introduces a Universal Marginaliser based on a neural network which can approximate all conditional marginal distributions of a PGM. 
We have shown that a UM can be used via a chain decomposition of the BN to approximate the joint posterior distribution, and thus the optimal proposal distribution for importance sampling. 
While this process is computationally intensive, a first-order approximation can be used requiring only a single evaluation of a UM per evidence set. 
We evaluated the UM on multiple datasets and also on a large medical PGM demonstrating that the UM significantly improves the efficiency of importance sampling.
The UM was trained offline using a large amount of generated training samples and for this reason, the model learned an effective representation for amortising the cost of inference.
This speed-up makes the UM (in combination with importance sampling) applicable for interactive applications that require a high performance on very large PGMs.
Furthermore, we have explored the use of the UM embeddings and we have 
shown that they can be used for tasks such as classification, clustering and interpretability of node relations. These UM embeddings make it possible to build more robust machine learning applications that rely on large generative models.

\clearpage
\bibliographystyle{iclr2019_conference}
\bibliography{refs}

\begin{thebibliography}{28}
\providecommand{\natexlab}[1]{#1}
\providecommand{\url}[1]{\texttt{#1}}
\expandafter\ifx\csname urlstyle\endcsname\relax
  \providecommand{\doi}[1]{doi: #1}\else
  \providecommand{\doi}{doi: \begingroup \urlstyle{rm}\Url}\fi

\bibitem[Baldi \& Sadowski(2014)Baldi and Sadowski]{baldi2014dropout}
Pierre Baldi and Peter Sadowski.
\newblock The dropout learning algorithm.
\newblock \emph{Artificial intelligence}, 210:\penalty0 78--122, 2014.

\bibitem[Cheng \& Druzdzel(2000)Cheng and Druzdzel]{cheng2000ais}
Jian Cheng and Marek~J. Druzdzel.
\newblock {AIS-BN}: An adaptive importance sampling algorithm for evidential
  reasoning in large {B}ayesian networks.
\newblock \emph{Journal of Artificial Intelligence Research}, 2000.

\bibitem[Fan et~al.(2008)Fan, Chang, Hsieh, Wang, and Lin]{fan2008liblinear}
Rong-En Fan, Kai-Wei Chang, Cho-Jui Hsieh, Xiang-Rui Wang, and Chih-Jen Lin.
\newblock Liblinear: A library for large linear classification.
\newblock \emph{Journal of machine learning research}, 9\penalty0
  (Aug):\penalty0 1871--1874, 2008.

\bibitem[Friedman(2004)]{friedman2004inferring}
Nir Friedman.
\newblock Inferring cellular networks using probabilistic graphical models.
\newblock \emph{Science}, 303\penalty0 (5659):\penalty0 799--805, 2004.

\bibitem[Germain et~al.(2015)Germain, Gregor, Murray, and
  Larochelle]{germain2015made}
Mathieu Germain, Karol Gregor, Iain Murray, and Hugo Larochelle.
\newblock {MADE}: masked autoencoder for distribution estimation.
\newblock In \emph{Proceedings of the 32nd International Conference on Machine
  Learning (ICML-15)}, pp.\  881--889, 2015.

\bibitem[Gershman \& Goodman(2014)Gershman and Goodman]{gershman2014amortized}
Samuel Gershman and Noah Goodman.
\newblock Amortized inference in probabilistic reasoning.
\newblock In \emph{Proceedings of the Cognitive Science Society}, volume~36,
  2014.

\bibitem[Gu et~al.(2015)Gu, Ghahramani, and Turner]{gu2015neural}
Shixiang Gu, Zoubin Ghahramani, and Richard~E Turner.
\newblock Neural adaptive sequential monte carlo.
\newblock In \emph{Advances in Neural Information Processing Systems}, pp.\
  2629--2637, 2015.

\bibitem[Hastings(1970)]{hastings1970monte}
W~Keith Hastings.
\newblock Monte carlo sampling methods using markov chains and their
  applications.
\newblock \emph{Biometrika}, 57\penalty0 (1):\penalty0 97--109, 1970.

\bibitem[Heckerman(1990)]{heckerman1990tractable}
David Heckerman.
\newblock A tractable inference algorithm for diagnosing multiple diseases.
\newblock In \emph{Machine Intelligence and Pattern Recognition}, volume~10,
  pp.\  163--171. Elsevier, 1990.

\bibitem[Heskes(2003)]{heskes2003stable}
Tom Heskes.
\newblock Stable fixed points of loopy belief propagation are local minima of
  the bethe free energy.
\newblock In \emph{Advances in neural information processing systems}, pp.\
  359--366, 2003.

\bibitem[Hornik et~al.(1989)Hornik, Stinchcombe, and White]{hornik}
Kurt Hornik, Maxwell Stinchcombe, and Halbert White.
\newblock Multilayer feedforward networks are universal approximators.
\newblock \emph{Neural networks}, 2\penalty0 (5):\penalty0 359--366, 1989.

\bibitem[Jaakkola \& Jordan(1999)Jaakkola and Jordan]{jaakkola1999variational}
Tommi~S Jaakkola and Michael~I Jordan.
\newblock Variational probabilistic inference and the {QMR-DT} network.
\newblock \emph{Journal of artificial intelligence research}, 10:\penalty0
  291--322, 1999.

\bibitem[Jordan et~al.(1999)Jordan, Ghahramani, Jaakkola, and
  Saul]{jordan1999introduction}
Michael~I Jordan, Zoubin Ghahramani, Tommi~S Jaakkola, and Lawrence~K Saul.
\newblock An introduction to variational methods for graphical models.
\newblock \emph{Machine learning}, 37\penalty0 (2):\penalty0 183--233, 1999.

\bibitem[Kingma \& Ba(2014)Kingma and Ba]{kingma2014adam}
Diederik Kingma and Jimmy Ba.
\newblock Adam: A method for stochastic optimization.
\newblock \emph{arXiv preprint arXiv:1412.6980}, 2014.

\bibitem[Koller \& Friedman(2009)Koller and Friedman]{koller2009probabilistic}
Daphne Koller and Nir Friedman.
\newblock \emph{Probabilistic graphical models: principles and techniques}.
\newblock MIT press, 2009.

\bibitem[Le et~al.(2017)Le, Baydin, Zinkov, and Wood]{le2017using}
Tuan~Anh Le, Atilim~Gunes Baydin, Robert Zinkov, and Frank Wood.
\newblock Using synthetic data to train neural networks is model-based
  reasoning.
\newblock \emph{arXiv preprint arXiv:1703.00868}, 2017.

\bibitem[Mnih \& Gregor(2014)Mnih and Gregor]{mnih2014neural}
Andriy Mnih and Karol Gregor.
\newblock Neural variational inference and learning in belief networks.
\newblock \emph{arXiv preprint arXiv:1402.0030}, 2014.

\bibitem[Morris(2001)]{Morris:2001:RNA:2074022.2074068}
Quaid Morris.
\newblock Recognition networks for approximate inference in bn20 networks.
\newblock In \emph{Proceedings of the Seventeenth Conference on Uncertainty in
  Artificial Intelligence}, UAI'01, pp.\  370--377, San Francisco, CA, USA,
  2001. Morgan Kaufmann Publishers Inc.
\newblock ISBN 1-55860-800-1.
\newblock URL \url{http://dl.acm.org/citation.cfm?id=2074022.2074068}.

\bibitem[Murphy et~al.(1999)Murphy, Weiss, and Jordan]{murphy1999loopy}
Kevin~P Murphy, Yair Weiss, and Michael~I Jordan.
\newblock Loopy belief propagation for approximate inference: An empirical
  study.
\newblock In \emph{Proceedings of the Fifteenth conference on Uncertainty in
  artificial intelligence}, pp.\  467--475. Morgan Kaufmann Publishers Inc.,
  1999.

\bibitem[Neal(2001)]{neal2001annealed}
Radford~M Neal.
\newblock Annealed importance sampling.
\newblock \emph{Statistics and computing}, 11\penalty0 (2):\penalty0 125--139,
  2001.

\bibitem[Ng \& Jordan(2000)Ng and Jordan]{ng2000approximate}
Andrew~Y Ng and Michael~I Jordan.
\newblock Approximate inference algorithms for two-layer bayesian networks.
\newblock In \emph{Advances in neural information processing systems}, pp.\
  533--539, 2000.

\bibitem[Paige \& Wood(2016)Paige and Wood]{paige2016inference}
Brooks Paige and Frank Wood.
\newblock Inference networks for sequential {M}onte {C}arlo in graphical
  models.
\newblock In \emph{International Conference on Machine Learning}, pp.\
  3040--3049, 2016.

\bibitem[Rezende et~al.(2014)Rezende, Mohamed, and
  Wierstra]{rezende2014stochastic}
Danilo~Jimenez Rezende, Shakir Mohamed, and Daan Wierstra.
\newblock Stochastic backpropagation and approximate inference in deep
  generative models.
\newblock \emph{arXiv preprint arXiv:1401.4082}, 2014.

\bibitem[Ritchie et~al.(2016)Ritchie, Horsfall, and Goodman]{ritchie2016deep}
Daniel Ritchie, Paul Horsfall, and Noah~D Goodman.
\newblock Deep amortized inference for probabilistic programs.
\newblock \emph{arXiv preprint arXiv:1610.05735}, 2016.

\bibitem[Shwe \& Cooper(1991)Shwe and Cooper]{shwe1991empirical}
Michael Shwe and Gregory Cooper.
\newblock An empirical analysis of likelihood-weighting simulation on a large,
  multiply connected medical belief network.
\newblock \emph{Computers and Biomedical Research}, 24\penalty0 (5):\penalty0
  453--475, 1991.

\bibitem[Srivastava et~al.(2014)Srivastava, Hinton, Krizhevsky, Sutskever, and
  Salakhutdinov]{JMLR:v15:srivastava14a}
Nitish Srivastava, Geoffrey Hinton, Alex Krizhevsky, Ilya Sutskever, and Ruslan
  Salakhutdinov.
\newblock Dropout: A simple way to prevent neural networks from overfitting.
\newblock \emph{Journal of Machine Learning Research}, 15:\penalty0 1929--1958,
  2014.
\newblock URL \url{http://jmlr.org/papers/v15/srivastava14a.html}.

\bibitem[Vincent et~al.(2008)Vincent, Larochelle, Bengio, and
  Manzagol]{vincent2008extracting}
Pascal Vincent, Hugo Larochelle, Yoshua Bengio, and Pierre-Antoine Manzagol.
\newblock Extracting and composing robust features with denoising autoencoders.
\newblock In \emph{Proceedings of the 25th international conference on Machine
  learning}, pp.\  1096--1103. ACM, 2008.

\bibitem[Wainwright et~al.(2008)Wainwright, Jordan,
  et~al.]{wainwright2008graphical}
Martin~J Wainwright, Michael~I Jordan, et~al.
\newblock Graphical models, exponential families, and variational inference.
\newblock \emph{Foundations and Trends{\textregistered} in Machine Learning},
  1\penalty0 (1--2):\penalty0 1--305, 2008.

\end{thebibliography}

\clearpage
\onecolumn
\section{Appendices}
\label{sec:appendix}
\label{sec:app:impo-sampling}
In this section, we review IS and describe how it is used for computing the marginals of a PGM given a set of evidence.

\subsection{Sampling with a Proposal Distribution}
\label{ap:is}

In BN inference, Importance Sampling (IS) is used to provide the posterior marginal estimates $P(\XU|\XO)$. To do so, we draw samples $\xU$ from a distribution $Q(\XU|\XO)$, known as the \textit{proposal} distribution. The proposal distribution must be defined such that we can both sample from and evaluate it efficiently. Provided we can evaluate $P(\XU,\XO)$, and that this distribution is such that $\XU$ contain the Markov boundary of $\XO$ along with all its ancestors, IS states that we can form posterior estimates: 
  \begin{equation}
    \label{eq:importance-sampling}
    \begin{aligned}
      P(\XU=\xU|\XO=\xO)
      &=\int 1_{\xU}(\mathbf x) \frac{P(\mathbf x|\xO)}{Q(\mathbf x| \xO)}Q(\mathbf x| \xO)d \mathbf x\\
      &=\frac{Q(\xO)}{P(\xO)}\int 1_{\xU}(\mathbf x) \frac{P(\mathbf x,\xO)}{Q(\mathbf x, \xO)}Q(\mathbf x| \xO)d \mathbf x\\
      &=\lim_{n\to\infty} \sum_{i=1}^{n} 1_{\xU}(\mathbf x_i) \frac{w_i}{\sum_{j=1}^{n} w_j},\
    \end{aligned}
  \end{equation}

where $\mathbf x_i\sim Q$ and $w_i=P(\mathbf x_i, \xO)/Q(\mathbf x_i, \xO)$ are the \textit{importance sampling weights} and $1_{\xU}(\mathbf x)$ is an indicator function for $\xU$. 

The simplest proposal distribution is the prior, $P(\XU)$.  However, as the prior and the posterior may be very different (especially in large networks) this is often an inefficient approach. An alternative is to use an estimate of the posterior distribution as a proposal. In this work, we argue that the UM learns an optimal proposal distribution.

\subsection{Sampling from the Posterior Marginals}
\label{ap:post_sample}
Take a BN with Bernoulli nodes and of arbitrary size and shape. Consider 2
specific nodes, $X_i$ and $X_j$, such that $X_j$ is caused only and always by
$X_i$:
\begin{align*}
  P(X_j=1|X_i=1) &= 1,\\
  P(X_j=1|X_i=0) &= 0.
\end{align*}
Given evidence $E$, we assume that $P(X_i|E)=0.001=P(X_j|E)$. We will now
illustrate that using the posterior distribution $P(X|E)$ as a proposal will not
necessarily yield the best result.

Say we have been given evidence, $E$, and the true conditional probability of $P(X_i|E) = 0.001$, therefore also $P(X_j|E) = 0.001$. We naively would expect $P(X|E)$ to be the optimal proposal distribution. However we can illustrate
the problems here by sampling with $Q=P(X|E)$ as the proposal.

Each node k $\in$ N will have a weight $w_k = P(X_k)/Q(X_k)$ and the total
weight of the sample will be $$w = \prod_{k=0}^{N} w_k.$$ The weights should be approximately 1 if Q is close to P. However, consider the $w_j$. There are four combinations of $X_i$ and $X_j$. We will sample $X_i$=1, $X_j$=1 only, in
expectation, one every million samples, however when we do the weight $w_j$ will be $w_j = P(X_j=1)/Q(X_j=1) = 1/0.001=1000$. This is not a problem in the limit, however if it happens for example in the first 1000 samples then it will
outweigh all other samples so far. As soon as we have a network with many nodes whose conditional probabilities are much greater than their marginal proposals this becomes almost inevitable. A further consequence of these high weights is that, since the entire sample is weighted by the same weight, every node probability will be effected by this high variance.
\section{Performance on large graphical models}
\newcommand{\WWW}{0.4}
\begin{figure}[ht!]
\hfill
\begin{subfigure}{\WWW\linewidth}
        \includegraphics[width=\linewidth]{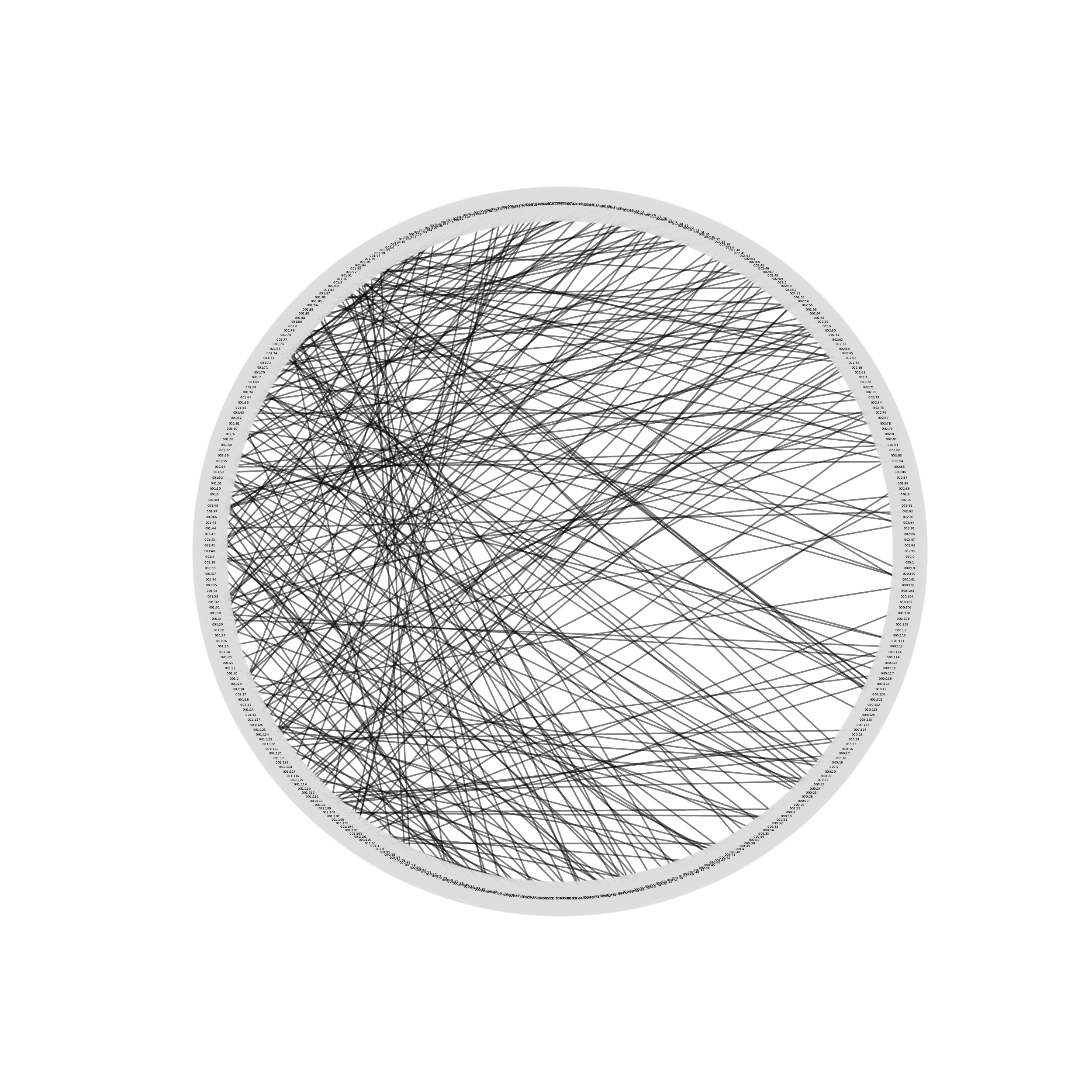}
        \includegraphics[width=\linewidth]{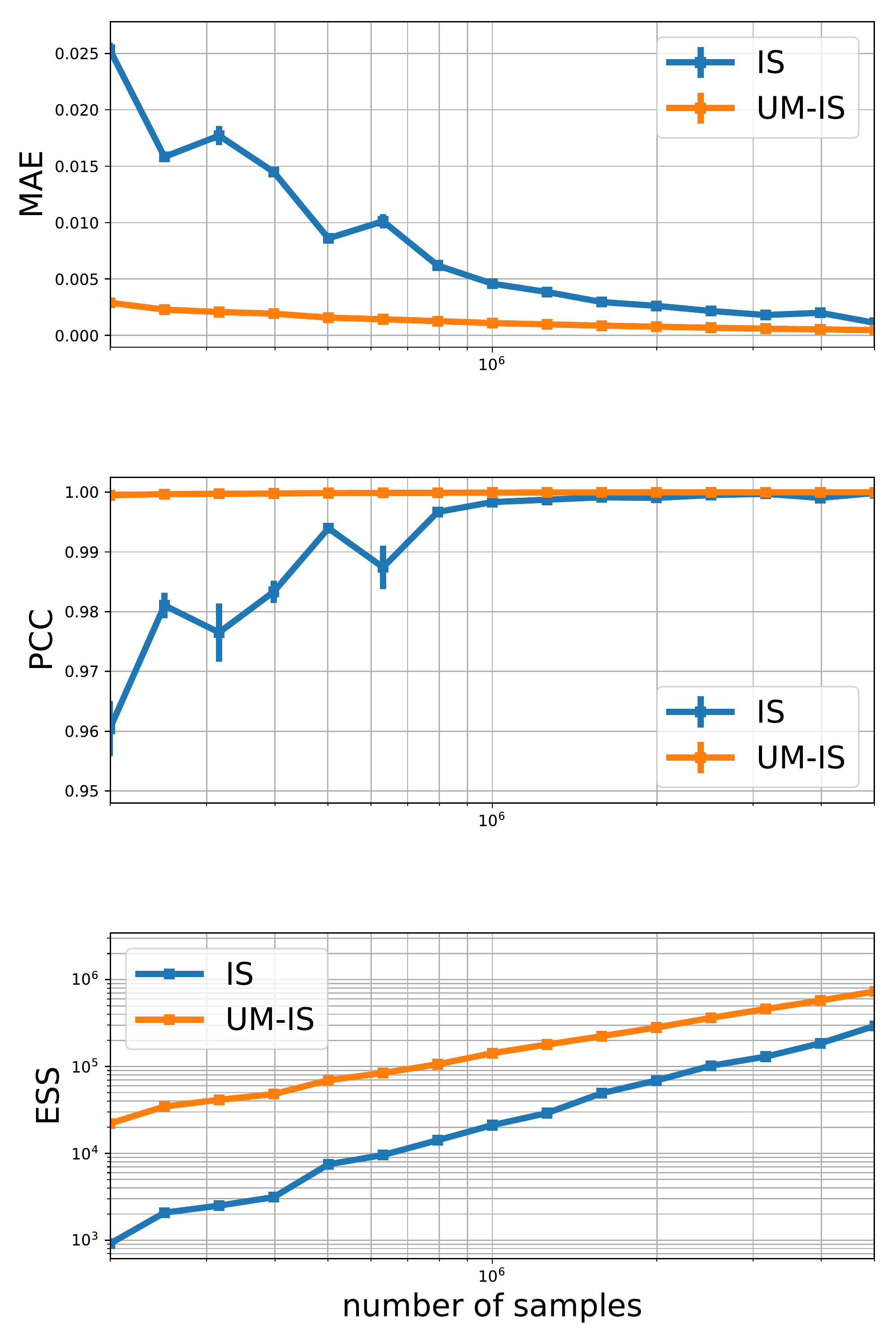}
        \caption{Synthetic graph, 384 nodes.}
\end{subfigure}
\hfill
\begin{subfigure}{\WWW\linewidth}
        \includegraphics[width=\linewidth]{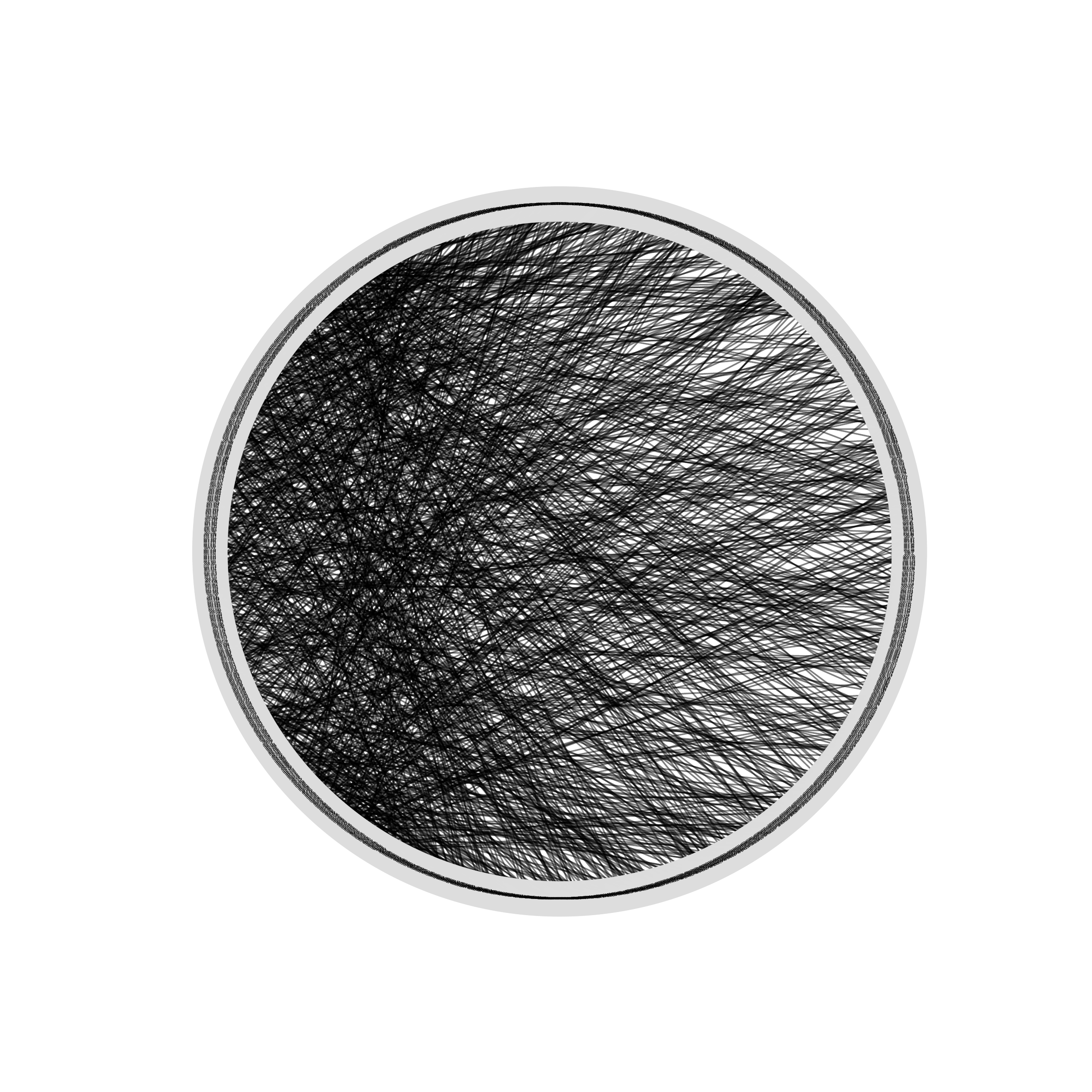}
        \includegraphics[width=\linewidth]{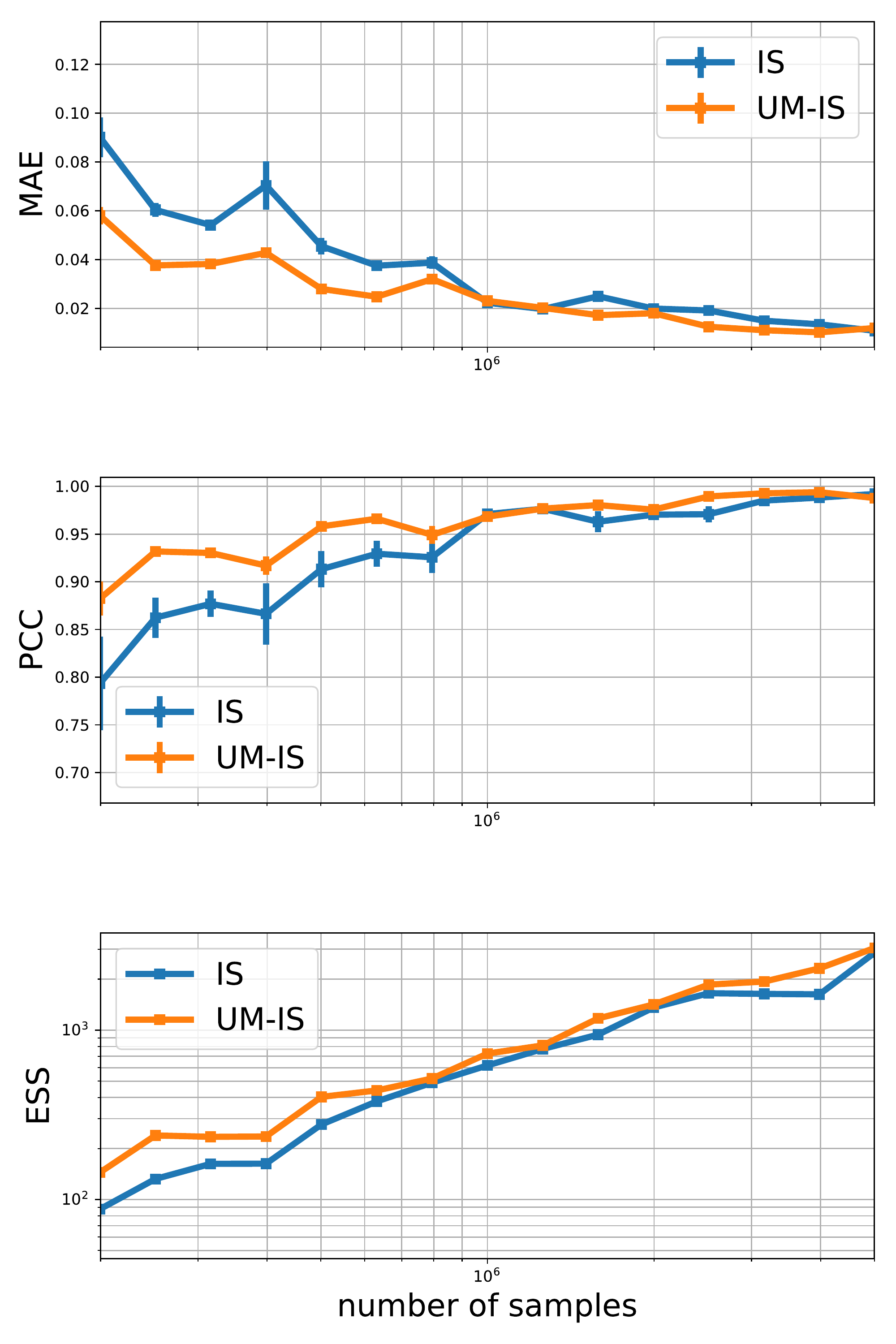}
        \caption{Synthetic graph, 1536 nodes.}
\end{subfigure}
        \caption{Additional experimetns on very large synthetic graphs.}
\label{fig:results2}
\hfill
\end{figure}

\section{Node Classification with UM Embedding}
\label{sec:node_classification}
{\bf  UM for Node Classification:}
The UM incorporates a encoding step, encoding the input layer into one shared Embedding layer (as discussed in Section \ref{sec:results}). To asses the quality of this embedding step, we perform classification experiments. First, we train a classifier from the input layer $S$ to the output layer $P$. Then, we compare to a classifier trained on the dense shared embedding to the output layer $P$. The dense shared embedding should encode all information present in the input layers separated for prediction (compare Fig.~\ref{fig:graph_embedding}). We compute embeddings for $853$ samples with $14$ diseases and use them to train an SVM classifier~\citep{fan2008liblinear} for disease detection.

See the comparison of classifier performance in Table \ref{tab:classification_embedding}. The learnt embedding significantly increases the performance for each classifier (about one order of magnitude). 

\begin{table}
    \centering
        \tiny
    \begin{tabular}{lcccccc}
    \toprule
    {} & \multicolumn{2}{c}{Linear SVC} & \multicolumn{2}{c}{RBF SVC} & \multicolumn{2}{c}{Ridge} \\
    {} &                            dense &          input &                            dense &          input &                            dense &          input \\
    \midrule
    F1        &                    $0.67\pm0.01$ &  $0.07\pm0.00$ &  $\mathbf{\mathbf{0.68\pm0.01}}$ &  $0.05\pm0.01$ &                    $0.66\pm0.04$ &  $0.17\pm0.01$ \\
    Precision &                    $0.84\pm0.03$ &  $0.20\pm0.04$ &  $\mathbf{\mathbf{0.85\pm0.02}}$ &  $0.16\pm0.09$ &                    $0.81\pm0.06$ &  $0.22\pm0.04$ \\
    Recall    &                    $0.58\pm0.02$ &  $0.05\pm0.00$ &                    $0.58\pm0.02$ &  $0.04\pm0.00$ &  $\mathbf{\mathbf{0.59\pm0.04}}$ &  $0.16\pm0.01$ \\
    Accuracy  &  $\mathbf{\mathbf{0.69\pm0.01}}$ &  $0.31\pm0.01$ &           $\mathbf{0.69\pm0.01}$ &  $0.32\pm0.01$ &                    $0.63\pm0.02$ &  $0.27\pm0.01$ \\
    \bottomrule
    \end{tabular}
    \caption{Classifier performances for three different classifiers. Each classifier is trained on - "dense" the dense embedding as features, and "input" - the top layer (UM input) as features.
    The target (output) is always the disease layer.}
    \label{tab:classification_embedding}
\end{table}

\end{document}